\documentclass[conference]{IEEEtran}
\IEEEoverridecommandlockouts
\usepackage{cite}
\usepackage{amsmath,amssymb,amsfonts}
\usepackage{algorithmic}
\usepackage{graphicx}
\usepackage{textcomp}
\usepackage{xcolor}
\usepackage{multirow}
\usepackage{bbding}
\usepackage{epstopdf}
\def\BibTeX{{\rm B\kern-.05em{\sc i\kern-.025em b}\kern-.08em
    T\kern-.1667em\lower.7ex\hbox{E}\kern-.125emX}}

\DeclareRobustCommand*{\IEEEauthorrefmark}[1]{%
    \raisebox{0pt}[0pt][0pt]{\textsuperscript{\footnotesize\ensuremath{#1}}}}

\renewcommand{\footnoterule}{%
  \kern -3pt 
  \hrule width 2in height 0.4pt 
  \kern 2.6pt 
}
    
\begin{document}

\title{Multimodal Mixture of Low-Rank Experts for Sentiment Analysis and Emotion Recognition}

\author{
    Shuo Zhang\IEEEauthorrefmark{1},
    Jinsong Zhang\IEEEauthorrefmark{2},
    Zhejun Zhang\IEEEauthorrefmark{1},
    Lei Li\IEEEauthorrefmark{1*}\thanks{*Corresponding author} \\
    \IEEEauthorblockA{\IEEEauthorrefmark{1}School of Artificial Intelligence, Beijing University of Posts and Telecommunications, Beijing, China} 
    \IEEEauthorblockA{\IEEEauthorrefmark{2}School of Computer Science and Technology, Harbin Institute of Technology, Weihai, China}
    \IEEEauthorblockA{Email: \{shuoz, zhejun.zhang, leili\}@bupt.edu.cn, 2024211124@stu.hit.edu.cn}
}

\maketitle

\begin{abstract}
Multi-task learning (MTL) enables the efficient transfer of extra knowledge acquired from other tasks. The high correlation between multimodal sentiment analysis (MSA) and multimodal emotion recognition (MER) supports their joint training. However, existing methods primarily employ hard parameter sharing, ignoring parameter conflicts caused by complex task correlations. In this paper, we present a novel MTL method for MSA and MER, termed \textit{Multimodal Mixture of Low-Rank Experts} (MMoLRE). MMoLRE utilizes shared and task-specific experts to distinctly model common and unique task characteristics, thereby avoiding parameter conflicts. Additionally, inspired by low-rank structures in the Mixture of Experts (MoE) framework, we design low-rank expert networks to reduce parameter and computational overhead as the number of experts increases. Extensive experiments on the CMU-MOSI and CMU-MOSEI benchmarks demonstrate that MMoLRE achieves state-of-the-art performance on the MSA task and competitive results on the MER task.
\end{abstract}


\section{Introduction}
Multimodal Sentiment Analysis (MSA) is a rapidly developing field that extends conventional text sentiment analysis by integrating text, audio, and visual signals. Recent advances in feature extraction and multimodal fusion have enabled a more nuanced understanding of human sentiment. At the same time, Akhtar et al. \cite{akhtar2019multi} proposed a novel approach by integrating Multimodal Emotion Recognition (MER) into a unified framework that jointly considers related tasks. Both MSA and MER utilize multimodal inputs to understand human affect, with MSA focusing on sentiment polarity or intensity and MER aiming to classify predefined emotional categories. UniMSE \cite{hu2022unimse} further embeds the MSA and MER tasks into the same feature space, leveraging the similarities and complementarity between sentiments and emotions to improve prediction. This multi-task learning (MTL) paradigm facilitates the seamless transfer of extra knowledge acquired from other tasks. However, existing methods that \textbf{straightforwardly share parameters between tasks may suffer from negative transfer} \cite{Tang2020Progressive} due to task conflicts. 

To address this issue, it is essential to separate shared and task-specific parameters to mitigate harmful interference between common and task-specific knowledge. One approach that has proven effective in this regard is the Mixture of Experts (MoE) \cite{Jacobs1991} model. Currently, MoE-based MTL methods \cite{Tang2020Progressive,Yang2024Multi} have achieved notable success. MoE can offer a unique advantage by dynamically encoding features for each sample and task, yielding diverse outputs with enhanced predictive power. In practice, expanding the number of experts can enhance the feature representation capacity and contextual understanding of the expert network, but it inevitably leads to higher model parameters and computational costs. 

Based on the above motivation, we propose a novel MTL framework called \textbf{M}ultimodal \textbf{M}ixture of \textbf{L}ow-\textbf{R}ank \textbf{E}xperts (MMoLRE). Specifically, to overcome the drawbacks of hard parameter sharing, we adopt the MoE architecture to construct task-specific features for each modality while assigning dedicated fusion networks for each task. Inspired by the use of low-rank structures in the MoE framework \cite{Yang2024Multi}, we transform the expert networks into a low-rank format to increase the number of experts without causing a substantial rise in parameters and FLOPs. Furthermore, our model is built upon state-of-the-art (SOTA) feature extraction and modality fusion techniques \cite{wu2024multimodal}. To verify the effectiveness of our method, we conduct comprehensive experiments on the CMU-MOSI and CMU-MOSEI datasets. 

In summary, our contributions are as follows:



\begin{itemize}
\item We propose a novel MTL framework, MMoLRE, which unifies the MSA and MER tasks to leverage their consensual and complementary information. Our approach explicitly separates shared and task-specific parameters to avoid parameter conflicts and achieve better performance.
\item The Unimodal Task-Specific Feature Extraction module in our model effectively captures the commonalities and specificities between the MSA and MER tasks using Mixture of Low-Rank Experts, while achieving over 80\% parameter savings compared to the standard MoE.
\item Experiments on the publicly popular CMU-MOSI and CMU-MOSEI datasets demonstrate that our method significantly outperforms previous SOTA approaches on the MSA task, while also exhibiting competitive performance on the MER task, despite not explicitly modeling contextual data.


\end{itemize}

\begin{figure*}[htbp]
\centering  
\includegraphics[width=0.835\linewidth]{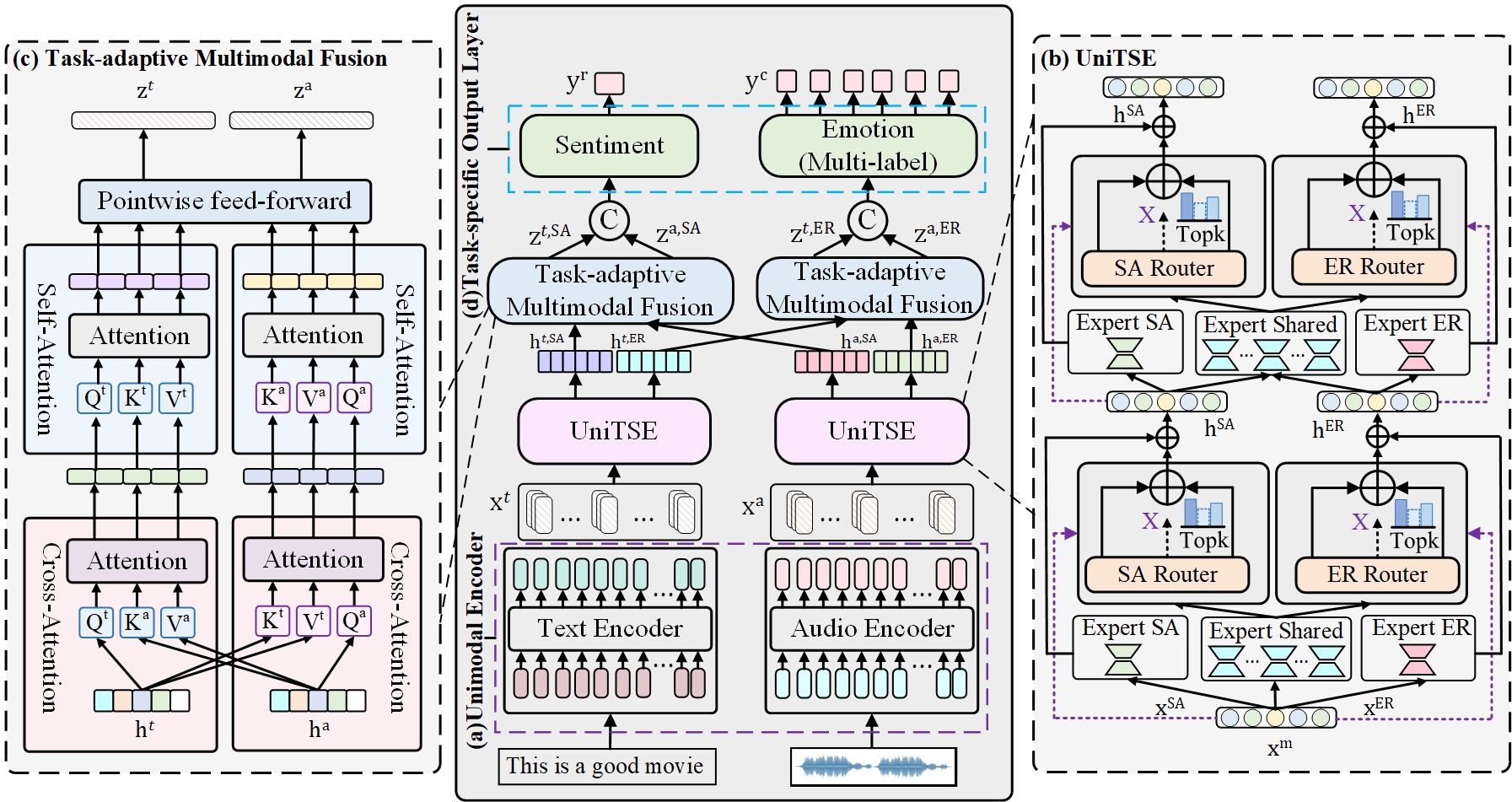}
\caption{The overview of MMoLRE.}
\label{fig:architecture}
\end{figure*}

\section{Related Work}
\subsection{Multimodal Sentiment Analysis and Emotion Recognition}
MSA approaches can be broadly divided into cross-modal sentiment semantic alignment and multimodal sentiment semantic fusion. In alignment-based methods, MulT \cite{Tsai2019multimodal} and HyCon \cite{Mai2023Hybrid} utilize cross-modal attention and contrastive learning, respectively, to achieve modality alignment. In contrast, fusion-based methods like MMML \cite{wu2024multimodal} first extract fine-grained representations for each modality at every time step and then integrate these multimodal features to capture both cross-modal and cross-temporal sentiment interactions. 

MER is similar to MSA, with the key distinction of focusing on identifying predefined emotions within conversations. Unlike MSA, which may concentrate on specific utterances, MER typically models multiple utterances to capture dynamic emotions. For example, COGMEN \cite{joshi-etal-2022-cogmen} employs graph neural networks to capture the inter- and intra-dependencies of utterances or speakers for multimodal emotion detection.

However, these studies treat MSA and MER as separate tasks, overlooking the potential of joint training. Integrating MSA and MER capitalizes on their complementary strengths, thereby improving model performance and generalization. The CMU-MOSEI \cite{bagher2018multimodal} dataset provides annotations for both tasks, and models like MTL \cite{akhtar2019multi} and UniMSE \cite{hu2022unimse} have introduced unified frameworks to address them collaboratively. Our research also emphasizes a unified framework, differing from previous methods by explicitly modeling the similarities and differences between tasks. 

\subsection{Multi-Task Learning}
MTL is a learning paradigm that effectively utilizes task-specific and shared information to tackle multiple related tasks simultaneously, and it has been extensively studied. Techniques such as cross-task attention \cite{Xu2023DeMT} enable networks to automatically integrate features learned from different tasks. However, scaling high-capacity models to adapt to multi-task settings remains a challenge. MoE \cite{Jacobs1991} offers a potential solution by leveraging multiple expert networks and a router network that determines each expert’s contribution to the final output. Building on this foundation, PLE \cite{Tang2020Progressive} explicitly separates shared experts to reduce interference between shared and task-specific knowledge. Mod-Squad \cite{Chen2023Mod-Squad} assigns experts to specific tasks by measuring the mutual information (MI) between tasks and experts. Unlike them, our approach employs a low-rank structure to reduce the parameters and computational overhead that scale with the number of experts.



\section{Methodology}
In this study, we concentrate exclusively on text and audio modalities. Fig.~\ref{fig:architecture} depicts the overall architecture of MMoLRE, which comprises four primary components: (a) the unimodal encoder, (b) the unimodal task-specific feature extraction (UniTSE) module, (c) the task-adaptive multimodal fusion module, and (d) the task-specific output layer. In the following sections, we formally define the tasks and provide a comprehensive description of each component.

\subsection{Notations and Task Definition}
Let the training dataset be \( D = \left\{ \left( I_{i}, y_{i}^{r}, y_{i}^{c} \right) \right\}_{i=1}^{O} \), where \( O \) is the number of utterances. Each utterance \( I_{i} = \left\{ I_{i}^{t}, I_{i}^{a} \right\} \) comprises raw unimodal sequences for text and audio modalities, with \( t \) and \( a \) representing text and audio, respectively. The label \( y_{i}^{r} \in \mathbb{R} \) denotes the sentiment intensity, while \( y_{i}^{c} \in \{0,1\}^{C} \) is a binary vector indicating the presence or absence of each of the \( C \) distinct emotion classes. Our goal is to simultaneously predict the sentiment intensity and the emotion categories from the multimodal data within each utterance.

For each utterance \( I_{i} \), we derive two independent representations, \( \mathbf{x}_{i}^{t} \in \mathbb{R}^{T_{t} \times d} \) and \( \mathbf{x}_{i}^{a} \in \mathbb{R}^{T_{a} \times d} \), obtained from RoBERTa \cite{liu2019roberta} and Data2Vec \cite{baevski2022data2vec}, respectively. Here, \( T_{t} \) and \( T_{a} \) represent the sequence lengths of the text and audio features, respectively, and \( d \) denotes the dimensionality of the feature representations, where  \( d=768 \). 

\subsection{Unimodal Task-Specific Feature Extraction}
\label{section3.2}
The UniTSE module is central to MMoLRE, as illustrated in Fig.~\ref{fig:architecture}(b). It comprises two identical blocks stacked sequentially, each serving distinct purposes to enable effective feature extraction across different tasks. The first block directly processes the features from the encoder and separates the task-specific features, while the second block facilitates the modeling of inter-task correlations through shared processing.

Initially, features from the encoder are duplicated for each task, resulting in \( \mathbf{x}_{i}^{\text{SA}} = \mathbf{x}_{i}^{\text{ER}} = \mathbf{x}_{i}^{m} \), where \( m \in \{t, a\} \) indicates the modality. For each task \( \tau \in \{\text{SA}, \text{ER}\} \), the task-specific features \( \mathbf{x}_{i}^{\tau} \) are processed through three pathways: a set of shared low-rank experts \( \{f_{\text{sh}}^{(n)}(\cdot)\}_{n=1}^{N} \), a task-specific low-rank expert \( f_{\text{ts}}^{\tau}(\cdot) \), and a task-specific router network \( g_{\tau}(\cdot) \). 

Each shared expert consists of a sequence of 1D convolutional layers with kernel sizes of 3 and 1. The weights and biases of the shared experts are denoted as \( \{w_{\text{sh1}}^{(n)}, b_{\text{sh1}}^{(n)}, w_{\text{sh2}}^{(n)}, b_{\text{sh2}}^{(n)}\} \) for \( n \in \{1, \dots, N\} \). Specifically, \( w_{\text{sh1}}^{(n)} \in \mathbb{R}^{3 \times d \times r_n} \), \( b_{\text{sh1}}^{(n)} \in \mathbb{R}^{r_n} \), \( w_{\text{sh2}}^{(n)} \in \mathbb{R}^{3 \times r_n \times d} \), and \( b_{\text{sh2}}^{(n)} \in \mathbb{R}^{d} \), where \( r_n \ll d \) ensures low-rank approximations. This configuration allows each shared expert to effectively capture the nuances of specific patterns while maintaining parameter efficiency. Similarly, the task-specific experts \( f_{\text{ts}}^{\tau}(\cdot) \) are designed with an analogous structure to the shared experts. Their parameters are defined as \( w_{\text{ts1}}^{\tau} \in \mathbb{R}^{3 \times d \times r_{\tau}} \), \( b_{\text{ts1}}^{\tau} \in \mathbb{R}^{r_{\tau}} \), \( w_{\text{ts2}}^{\tau} \in \mathbb{R}^{3 \times r_{\tau} \times d} \), and \( b_{\text{ts2}}^{\tau} \in \mathbb{R}^{d} \), where \( r_{\tau} \ll d \). The task-specific experts enhance the saliency of task-specific features, thereby improving the model's ability to distinguish and prioritize relevant information for each individual task.

The router network \( g_{\tau}(\cdot) \) is responsible for dynamically selecting the most relevant shared experts based on the input features. Specifically, for each task \( \tau \), the router processes \( \mathbf{x}_{i}^{\tau} \) through two consecutive 1D convolutional layers with a kernel size of 1, reducing the dimension from \( d \) to \( d/4 \). This is followed by a global average pooling layer, and a softmax function is applied to generate gating values \( g_{\tau,n} \) for each shared expert \( n \). The top-\( k \) experts with the highest gating values are then selected, and their outputs are aggregated to form the task-specific feature \( \mathbf{h}_{i}^{\tau} \). Mathematically, this is expressed as:
\begin{equation}
    \mathbf{h}_{i}^{\tau} = \sum_{n \in \text{top-}k} g_{\tau,n} \cdot f_{\text{sh}}^{(n)}(\mathbf{x}_{i}^{\tau}) + f_{\text{ts}}^{\tau}(\mathbf{x}_{i}^{\tau}).
\end{equation} 

This architecture enables the modeling of inter-task correlations through shared processing while permitting task-specific adaptations. Notably, with the number of shared experts at 15, the adoption of a low-rank structure results in a reduction of parameters and FLOPs by 5.7$\times$ and 5.9$\times$, respectively, compared to the standard MoE. Consequently, the UniTSE module effectively balances parameter efficiency with the flexibility required to handle multiple tasks within the multimodal framework.  

Furthermore, our UniTSE module exhibits structural similarities to the Progressive Layered Extraction (PLE) \cite{Tang2020Progressive} model, but it omits the gating network for shared experts. Instead, it benefits from early feature separation, fully leveraging the high-quality features generated by the powerful encoders and the significant correlation between sentiment analysis and emotion recognition tasks. 

\subsection{Task-Adaptive Multimodal Fusion} 
The Task-Adaptive Multimodal Fusion module integrates the task-specific features from both modalities to form a unified representation. This fusion effectively captures the complementary information inherent in each modality.
    
For each task \( \tau \), let \( \mathbf{h}_{i}^{t, \tau} \) and \( \mathbf{h}_{i}^{a, \tau} \) denote the task-specific features for the text and audio modalities, respectively, obtained from the UniTSE module. The fusion process comprises three key components. First, cross-attention mechanisms are employed to capture the interdependencies between the text and audio modalities. Specifically, features from one modality are used to query the other modality, enabling each modality to attend to the relevant information from its counterpart. Formally, the cross-attention operations are defined as:
\begin{align}
    \mathbf{z}_{i}^{t, \tau} &= \text{Attention}(\mathbf{W}_q \mathbf{h}_{i}^{t, \tau}, \mathbf{W}_k \mathbf{h}_{i}^{a, \tau}, \mathbf{W}_v \mathbf{h}_{i}^{a, \tau}), \\
    \mathbf{z}_{i}^{a, \tau} &= \text{Attention}(\mathbf{W}_q \mathbf{h}_{i}^{a, \tau}, \mathbf{W}_k \mathbf{h}_{i}^{t, \tau}, \mathbf{W}_v \mathbf{h}_{i}^{t, \tau}),
\end{align}
where \( \mathbf{W}_q \), \( \mathbf{W}_k \), and \( \mathbf{W}_v \) are learnable projection matrices that transform the input features into queries, keys, and values, respectively.
The attention function is defined as:
\begin{equation}
    \operatorname{Attention}(Q, K, V) = \operatorname{softmax}\left( \frac{Q K^\top}{\sqrt{d}} \right) V.
\end{equation} 

After cross-modality encoding, self-attention mechanisms are independently applied to the feature representations of each modality. This process enables each modality to refine its internal contextual representations by attending to its respective temporal information. Subsequently, a pointwise feed-forward network, comprising fully connected layers and ReLU activation functions applied to each individual position, further enhances the encoded feature representations.

Upon processing through these components across \( L \) layers, the \texttt{[cls]} tokens from each modality are concatenated to construct a comprehensive fused representation. 

\begin{table*}[t]
\caption{Results of the MSA task on CMU-MOSI and CMU-MOSEI. All metrics are averaged over three runs. The best and second-best results are indicated by bold and underlined text, respectively.}
\label{tab:result1}
\large
\resizebox{\linewidth}{!}{
\begin{tabular}{c|cccccccc|cccccccc}
\hline
\multirow{2}{*}{\textbf{Method}} & \multicolumn{8}{c|}{\textbf{CMU-MOSI}} & \multicolumn{8}{c}{\textbf{CMU-MOSEI}} \\
& \textbf{Acc2$_{Has0}\uparrow$} & \textbf{F1$_{Has0}\uparrow$} & \textbf{Acc2$_{Non0}\uparrow$} & \textbf{F1$_{Non0}\uparrow$} & \textbf{Acc5$\uparrow$} & \textbf{Acc7$\uparrow$} & \textbf{MAE$\downarrow$} & \textbf{Corr$\uparrow$} & \textbf{Acc2$_{Has0}\uparrow$} & \textbf{F1$_{Has0}\uparrow$} & \textbf{Acc2$_{Non0}\uparrow$} & \textbf{F1$_{Non0}\uparrow$} & \textbf{Acc5$\uparrow$} & \textbf{Acc7$\uparrow$} & \textbf{MAE$\downarrow$} & \textbf{Corr$\uparrow$} \\
\hline
MTL & - & - & - & - & - & - & - & - & 78.8 & 80.5 & - & - & - & - & - & - \\
MulT & 81.50 & 80.60 & 84.10 & 83.90 & - & - & 0.861 & 0.711 & - & - & 82.5 & 82.3 & - & - & 0.580 & 0.713 \\
MISA & 80.79 & 80.77 & 82.10 & 82.03 & - & - & 0.804 & 0.764 & 82.59 & 82.67 & 84.23 & 83.97 & - & - & 0.568 & 0.717 \\
Self-MM & 84.00 & 84.42 & 85.98 & 85.95 & - & - & 0.713 & 0.798 & 82.81 & 82.53 & 85.17 & 85.30 & - & - & 0.530 & 0.765 \\
MAG-BERT & 84.20 & 84.10 & 86.10 & 86.00 & - & - & 0.712 & 0.796 & 84.70 & 84.50 & - & - & - & - & - & - \\
MMIM & 84.14 & 84.00 & 86.06 & 85.98 & - & 46.65 & 0.700 & 0.800 & 82.24 & 82.66 & 85.97 & 85.94 & - & 54.24 & 0.526 & 0.772 \\
SPECTRA & - & - & 87.5 & - & - & - & - & - & - & - & 87.34 & - & - & - & - & - \\
HyCon & - & - & 86.4 & 86.4 & - & \underline{48.3} & \underline{0.664} & \underline{0.832} & - & - & 86.5 & 86.4 & - & 53.4 & 0.590 & 0.792 \\
UniMSE & \underline{85.85} & \underline{85.83} & 86.9 & 86.42 & - & \textbf{48.68} & 0.691 & 0.809 & \underline{85.86} & \underline{85.79} & \textbf{87.5} & \textbf{87.46} & - & \underline{54.39} & \underline{0.523} & 0.773 \\
MMML & 85.57 & 85.52 & \underline{87.76} & \underline{87.76} & \underline{55.83} & 47.03 & \textbf{0.663} & \textbf{0.837} & 85.55 & 85.58 & \underline{86.73} & \underline{86.51} & \underline{56.79} & 54.38 & 0.526 & 0.79 \\
\hline
\textbf{MMoLRE} & \textbf{85.96} & \textbf{85.93} & \textbf{88.21} & \textbf{88.23} & \textbf{56.51} & 47.52 & 0.666 & \textbf{0.837} & \textbf{86.31} & \textbf{86.19} & 86.57 & 86.28 & \textbf{57.96} & \textbf{55.78} & \textbf{0.505} & \textbf{0.797} \\
\hline
\end{tabular}
}
\end{table*}

\begin{table*}
\caption{Results of the MER task on CMU-MOSEI.}
\label{tab:result2}
\resizebox{\linewidth}{!}{
\begin{tabular}{c|ccccccccccccccc}
\hline
\multirow{2}{*}{\textbf{Method}} & \multirow{2}{*}{\textbf{Context Modeling}} & \multicolumn{2}{c}{\textbf{Happiness}} & \multicolumn{2}{c}{\textbf{Sadness}}   & \multicolumn{2}{c}{\textbf{Anger}}     & \multicolumn{2}{c}{\textbf{Surprise}}  & \multicolumn{2}{c}{\textbf{Disgust}}   & \multicolumn{2}{c}{\textbf{Fear}}      & \multicolumn{2}{c}{\textbf{Average}}   \\
                        &                                   & \textbf{ACC}           & \textbf{WF1}           & \textbf{ACC}           & \textbf{WF1}           & \textbf{ACC}           & \textbf{WF1}           & \textbf{ACC}           & \textbf{WF1}           & \textbf{ACC}           & \textbf{WF1}           & \textbf{ACC}           & \textbf{WF1}           & \textbf{ACC}           & \textbf{WF1}           \\ \hline
Multilogue-Net                 & \Checkmark                        & -             & 70.8          & -             & 70.9          & -             & 74.5          & -             & 87.7          & -             & 83.6          & -             & 86.2          & -             & 79.0          \\
TBJE                    & \XSolidBrush                        & 66.0          & 65.5          & 73.9          & 67.9          & \textbf{81.9} & 76.0          & 90.6          & 86.1          & \textbf{86.5} & 84.5          & 89.2          & 87.2 & 81.4          & 77.9          \\
MR                    & \XSolidBrush                        & \underline{69.7}          & 69.4          & \textbf{76.0}          & 72.1          & 77.6 & 72.8          & \underline{91.7}          & \underline{87.8}          & 83.1 & 82.3          & \textbf{90.5}          & 86.0 & 81.4          & 78.4          \\
COGMEN                  & \Checkmark                        & -             & \underline{70.9} & -             & 70.9 & -             & 74.2 & -             & 86.1 & -             & 81.8 & -             & \underline{87.8}    & -             & 78.6   \\
CORECT                  & \Checkmark                        & -             & \textbf{71.4} & -             & \underline{72.9} & -             & \underline{76.8} & -             & 86.5 & -             & 84.3 & -             & \textbf{87.9}    & -             & \underline{80.0} 
\\ \hline
Baseline (only MER)            & \XSolidBrush                      & 68.2          & 68.0          & 73.0    & 71.9          & 76.4          & 75.6          & 88.7    & 87.7          & 84.1          & 84.0          & 88.0 & 84.0          & 79.7          & 78.9          \\
Pre-fusion             & \XSolidBrush                      & \textbf{70.3} & 70.3    & 73.7          & \textbf{73.6}    & \underline{79.7}    & \textbf{76.9}    & \textbf{91.8} & \textbf{88.0}    & 85.5          & \underline{85.2}          & \underline{90.2}    & 86.4          & \underline{81.9}    & \textbf{80.1}    \\
Post-fusion            & \XSolidBrush                      & 67.3          & 66.8          & 75.0    & 71.7          & 79.0          & 75.0          & \underline{91.7}    & \underline{87.8}          & 84.7          & 84.2          & \textbf{90.5} & 86.0          & 81.4          & 78.6          \\
MMoLRE (Pre-fusion+UniTSE)          & \XSolidBrush                      & 69.4    & 69.5          & \underline{75.8} & 72.5          & 79.5          & 75.6          & \underline{91.7}    & \underline{87.8}          & \underline{86.4}    & \textbf{85.4}    & \textbf{90.5} & 86.0          & \textbf{82.2} & 79.5          \\ \hline
\end{tabular}
}
\end{table*}

\subsection{Predictions and Losses}  
The task-specific output layer generates predictions for sentiment analysis and emotion classification through two distinct branches. Sentiment prediction is treated as a regression task, with the Mean Absolute Error (MAE) loss applied, defined as:
\begin{equation}
    \mathcal{L}_{\text{MAE}} = \frac{1}{B} \sum_{i=1}^{B} \left| y_i^r - \hat{y}_i^r \right|,
\end{equation}
where \( \hat{y}_i^r \) is the predicted value and \( B \) is the number of utterances. 
Emotion prediction is formulated as a multi-label classification task, with the cross-entropy loss used:
\begin{equation}
    \mathcal{L}_{\text{CE}} = - \frac{1}{B} \sum_{i=1}^{B} \sum_{j=1}^{C} y_{i,j}^{c} \log(\hat{y}_{i,j}^{c}),
\end{equation}
where \( y_{i,j}^{c} \) and \( \hat{y}_{i,j}^{c} \) denote the true label and predicted probability for class \( j \) of the \( i \)-th sample, respectively. 

In the multi-task learning setup, the final objective is the sum of these losses, optimized equally to balance sentiment analysis and emotion classification. 

\section{Experiments}

\subsection{Experimental Setup}

\subsubsection{Datasets} We evaluate our MMoLRE on two public multimodal sentiment datasets. CMU-MOSI \cite{Zadeh2016Multimodal} consists of 2,199 video segments, each annotated with sentiment scores ranging from -3 to +3 to indicate sentiment polarity and intensity. CMU-MOSEI \cite{bagher2018multimodal}, an extension of CMU-MOSI, contains 22,856 video segments annotated with both sentiment and emotion labels. The emotion labels include six standard types: \textit{happiness}, \textit{sadness}, \textit{anger}, \textit{surprise}, \textit{disgust}, and \textit{fear}, each associated with a non-negative value that represents the degree of the emotion. However, most existing studies primarily focus on its sentiment annotations.

\subsubsection{Baselines} We compare the proposed method with various baseline models on both MSA and MER tasks. The baselines are categorized into single-task MSA methods (MulT \cite{Tsai2019multimodal}, MISA \cite{Hazarika2020MISA}, MAG-BERT \cite{rahman2020integrating}, Self-MM \cite{yu2021learning}, MMIM \cite{han2021improving}, SPECTRA \cite{yu2023speech}, HyCon \cite{Mai2023Hybrid}, MMML \cite{wu2024multimodal}), single-task MER methods (Multilogue-Net \cite{shenoy-sardana-2020-multilogue}, TBJE \cite{delbrouck-etal-2020-transformer},  MR \cite{tsai-etal-2020-multimodal}, COGMEN \cite{joshi-etal-2022-cogmen}, CORECT \cite{nguyen-etal-2023-conversation}), and multi-task learning approaches (MTL \cite{akhtar2019multi}, UniMSE \cite{hu2022unimse}).

\begin{figure*}[t]
\centering
\includegraphics[width=0.82\linewidth]{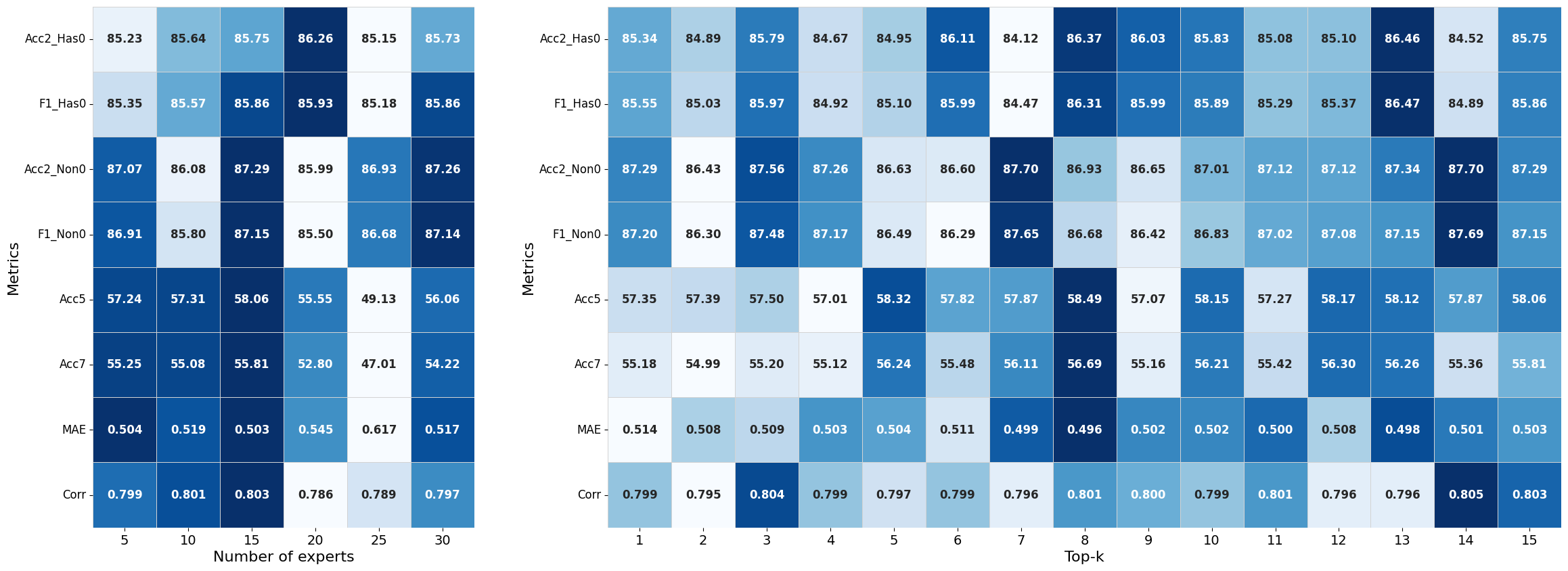}
\caption{Analysis of the number of shared low-rank experts and the top-$k$ selection of experts by the task-specific router networks on CMU-MOSEI. Each metric is mapped to the same numerical scale, with darker colors indicating better performance.}
\label{fig:analysis1}
\end{figure*}

\begin{figure}[t]
\centerline{\resizebox{0.5\textwidth}{!}{\includegraphics{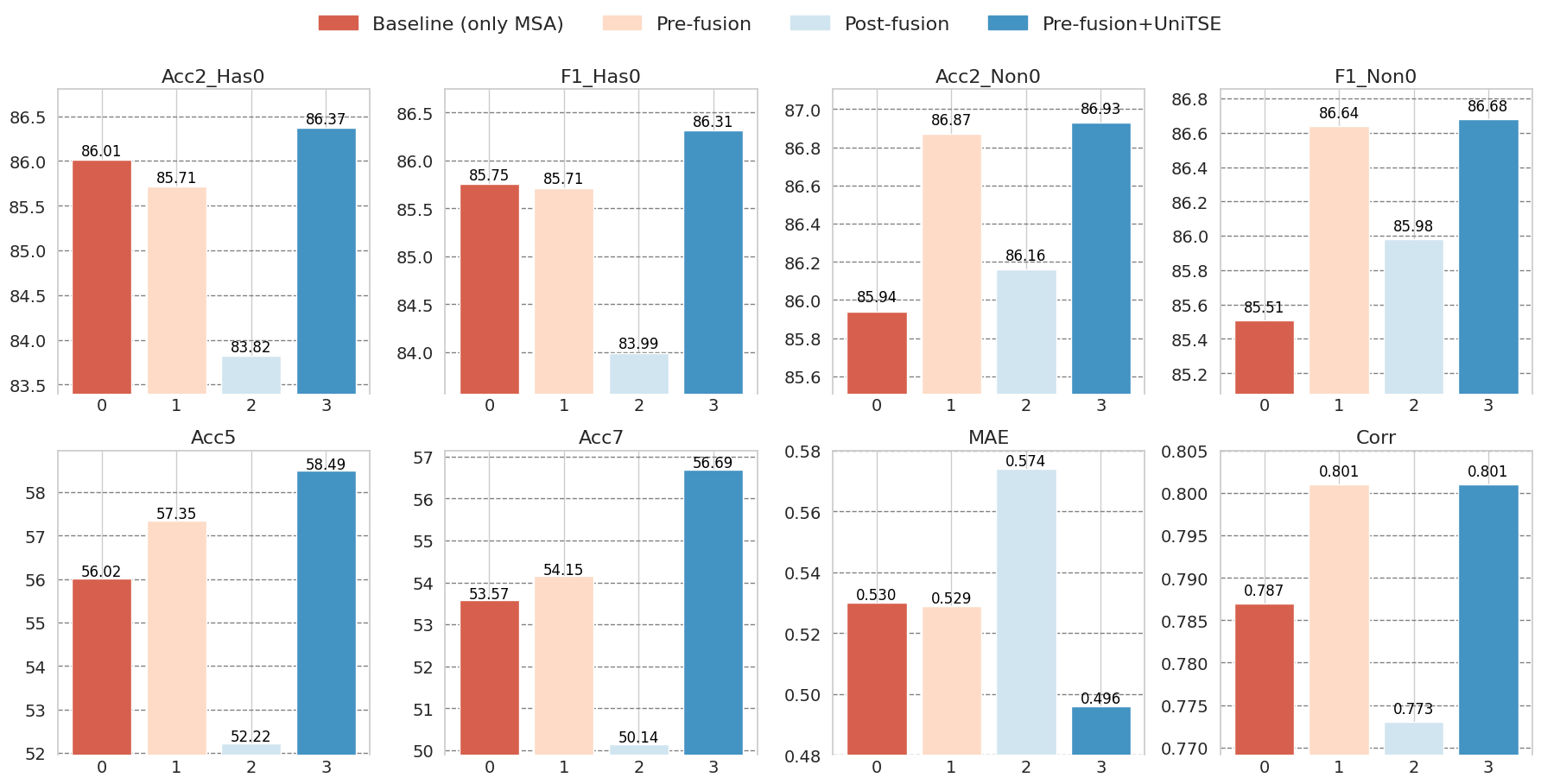}}}
\caption{Ablation study on CMU-MOSEI.}
\label{fig:ablation1}
\end{figure}
\subsubsection{Evaluation metrics} We evaluate our model using metrics consistent with prior studies. For MSA, we employ mean absolute error (MAE), Pearson correlation (Corr), seven-class classification accuracy (Acc7), five-class classification accuracy (Acc5), binary classification accuracy (Acc2), and F1 score. Specifically, for Acc2 and F1, we conduct two classification types: positive/negative classification excluding zero sentiment scores (Non0) and non-negative/negative classification including zero sentiment scores as positive (Has0). For MER, we utilize Accuracy and Weighted F1-score (WF1) as the evaluation metrics.



\subsection{Implementation} To address missing emotion labels in CMU-MOSI, we adopt the label unification method from UniMSE \cite{hu2022unimse}. Samples from CMU-MOSI and multiple emotion datasets (CMU-MOSEI, MELD \cite{poria-etal-2019-meld}, and IEMOCAP \cite{busso_iemocap_2008}) are first classified by sentiment polarity. We then compute the semantic similarity between samples with identical polarity across these datasets using SimCSE \cite{gao-etal-2021-simcse}. Based on these similarities, each CMU-MOSI sample is assigned emotion labels from the most similar sample among the combined emotion datasets. 

The model is trained with a learning rate of $1 \times 10^{-5}$, a batch size of 8, and the AdamW optimizer, incorporating early stopping with a patience of 8 epochs to enhance generalizability. We utilize base-sized pre-trained models for all modalities, keeping the CNN layers in the audio model frozen during feature extraction. Additionally, the rank of task-specific low-rank expert is set to 128, and the fusion network comprises 5 layers. For CMU-MOSI and CMU-MOSEI, the parameters are configured as $n=15$, $r_{n}=64$, and $k=11$ for CMU-MOSI, and $n=15$, $r_{n}=128$, and $k=8$ for CMU-MOSEI.

\subsection{Results}
We evaluate MMoLRE on the MSA and MER tasks using the CMU-MOSI and CMU-MOSEI datasets, presenting MSA results in Table~\ref{tab:result1} and MER results in Table~\ref{tab:result2}. 

For MSA, our model outperforms SOTA approaches on most evaluation metrics. On CMU-MOSEI, MMoLRE surpasses the MTL framework UniMSE \cite{hu2022unimse} in all metrics except Acc2$_{Non0}$ and F1$_{Non0}$, highlighting its potential to capture the distinctions between MSA and MER tasks. Compared to the previous SOTA, MMML \cite{wu2024multimodal}, we improve Acc5 and Acc7 by 1.17\% and 1.4\%, respectively, and reduce MAE by 0.021, demonstrating that incorporating the MER task can enhance MSA performance. Moreover, on CMU-MOSI, leveraging pseudo-labels for emotion classification leads to SOTA results, further validating the effectiveness of integrating the MER task.

For MER, our method does not explicitly model previous utterances, yet it still attains competitive performance. These findings highlight that learning MSA and MER simultaneously provides mutual benefits. 


\subsection{Ablation Study}
We conduct ablation experiments to validate the effectiveness of multi-task learning and the UniTSE module within our MMoLRE model. All experiments utilize the same feature extractor and modality fusion module. As shown in Fig.~\ref{fig:ablation1}, the quantitative results for the MSA task include a baseline model trained solely for the MSA task and two hard parameter-sharing MTL variants we designed, namely pre-fusion and post-fusion. In the pre-fusion setting, features are shared between tasks through the common feature extractor and then processed individually using task-specific modality fusion modules, differing from our proposed method by the absence of the UniTSE module. In the post-fusion setting, tasks additionally share the modality fusion module, and the fused features are passed through task-specific output layers for predictions. The results indicate that separating task features before modality fusion in MTL leads to better performance. Crucially, our proposed UniTSE module further improves model performance by effectively capturing the similarities and differences between the MSA and MER tasks, outperforming other configurations. Additionally, we report the accuracy and weighted F1 scores for the six emotion categories in the MER task in Table~\ref{tab:result2}, where similar conclusions are drawn. 


\begin{table}[tbp]
\caption{Analysis of the rank setting of each expert on CMU-MOSEI.}
\label{tab: analysis2}
\large
\begin{center}
\resizebox{\linewidth}{!}{
\begin{tabular}{c|cccccccc}
\hline
Rank Setting & Acc2$_{Has0}$ & F1$_{Has0}$ & Acc2$_{Non0}$ & F1$_{Non0}$ & Acc5 & Acc7 & MAE   & Corr  \\ \hline
$\left \{ 16 \right \} ^{15}$             & 86.18         & 86.11       & 86.30         & 86.06       & 57.76 & 55.23 & 0.512 & 0.802 \\
$\left \{ 64 \right \} ^{15}$             & 86.11         & 86.03       & 86.41         & 86.14       & 56.73 & 54.30 & 0.506 & 0.803 \\
$\left \{ 128 \right \} ^{15}$            & 85.75         & 85.86       & 87.29         & 87.15       & 58.06 & 55.81 & 0.503 & 0.803 \\
$\left \{ 16+8n | n=0,1,...,14 \right \}$ & 85.34         & 85.50       & 87.09         & 86.98       & 58.04 & 56.26 & 0.501 & 0.795 \\ \hline
\end{tabular}
}
\end{center}
\end{table}

\subsection{Parameter Sensitivity Study}
In this subsection, we conduct extensive experiments to explore the impact of three hyper-parameters on the performance of our MMoLRE model, including the number of shared low-rank experts, the rank setting of each expert, and the top-$k$ selection of experts by the task-specific router networks. Since our input settings are more suitable for MSA, we use the metrics from the MSA task as a reference. We fix two parameters and vary the remaining one to assess its effect on the MSA task. As shown in Fig.~\ref{fig:analysis1}, we initially set the rank of all shared experts to 128 and activate all of them. Under this configuration, we observe significant improvements as the number of experts increases, achieving peak performance when the number is set to 15. Next, we fix the number of experts at 15 and progressively reduce the number of activated experts. We find that setting $k=8$ offers the best trade-off across various metrics. However, activating too few experts leads to a substantial performance decline, which we hypothesize is due to the high correlation between the MSA and MER tasks, necessitating a more diverse set of experts.


Furthermore, with the number of shared experts fixed at 15 and all of them activated, we investigate the impact of different rank settings, including 1) all experts with a rank of 16, 2) all experts with a rank of 64, 3) all experts with a rank of 128, and 4) experts with ranks ranging from 16 to 128 in increments of 8. As presented in Table~\ref{tab: analysis2}, the third configuration achieves the best results. We analyze that lower-ranked experts may fail to adequately capture task-specific features, resulting in decreased overall model performance. 


\section{Conclusion}
In this paper, we propose a novel MTL model called Multimodal Mixture of Low-Rank Experts (MMoLRE), which jointly trains the MSA and MER tasks. By explicitly separating task-specific and task-common parameters, our model avoids parameter conflicts and effectively captures the similarities and differences between the tasks. Additionally, the low-rank expert structure efficiently controls parameter growth as the number of experts increases. Experimental results demonstrate that our proposed method outperforms previous SOTA approaches on the MSA task and achieves competitive performance on the MER task.

\section*{Acknowledgment}

This work was supported by the National Key R\&D Program of China under Grant 2024YFC3307802 and NSFC Grant No. 62176024.


\bibliographystyle{IEEEbib}
\bibliography{icme2025references}

\end{document}